# Brotherhood at WMT 2024: Leveraging LLM-Generated Contextual Conversations for Cross-Lingual Image Captioning

**Siddharth Betala**
betalas5@gmail.com

**Ishan Chokshi**
ishan.c1671@gmail.com

## Abstract

In this paper, we describe our system under the team name *Brotherhood* for the English-to-Lowres Multi-Modal Translation Task. We participate in the multi-modal translation tasks for English-Hindi, English-Hausa, English-Bengali, and English-Malayalam language pairs. We present a method leveraging multi-modal Large Language Models (LLMs), specifically GPT-4o and Claude 3.5 Sonnet, to enhance cross-lingual image captioning without traditional training or fine-tuning.

Our approach utilizes instruction-tuned prompting to generate rich, contextual conversations about cropped images, using their English captions as additional context. These synthetic conversations are then translated into the target languages. Finally, we employ a weighted prompting strategy, balancing the original English caption with the translated conversation to generate captions in the target language.

This method achieved competitive results, scoring 37.90 BLEU on the English-Hindi Challenge Set and ranking 1$^{st}$ and 2$^{nd}$ for English-Hausa on the Challenge and Evaluation Leaderboards, respectively. We conduct additional experiments on a subset of 250 images, exploring the trade-offs between BLEU scores and semantic similarity across various weighting schemes.

## 1 Introduction

Machine translation (MT) is a classic subfield in NLP that investigates the usage of computer software to translate text or speech from one language to another without human involvement (Yang et al., 2020). Machine translation (MT) has seen remarkable advancements in recent years, primarily due to the success of neural approaches (Bahdanau, 2014; Vaswani, 2017). However, these improvements have been predominantly observed in high-resource language pairs, leaving low-resource languages significantly behind (Sennrich and Zhang, 2019; Costa-jussà et al., 2022). The challenges in low-resource MT are multifaceted, including limited parallel corpora, lack of linguistic diversity in training data, and the absence of specialized tools and resources.

One promising direction to address these challenges is the incorporation of visual information into the translation process, known as multimodal machine translation (MMT) (Elliott et al., 2016; Specia et al., 2016; Calixto et al., 2017). The underlying hypothesis is that visual context can provide crucial disambiguating cues, especially for languages with limited textual resources. This approach aligns with the human cognitive process of language understanding, which often relies on multiple sensory inputs (Beinborn et al., 2018).

The Workshop on Machine Translation (WMT) 2024 has presented a shared task on English-to-Low-Resource Multi-Modal Translation, focusing on Hindi (Parida et al., 2019), Bengali (Sen et al., 2022), Malayalam[1], and Hausa (Abdulmumin et al., 2022). This task utilizes variants of the Visual Genome (Krishna et al., 2017) dataset, adapted for these target languages. While these datasets provide a valuable resource for research, they also present unique challenges. First, the quality of translations in low-resource languages can be inconsistent (see Table 1), potentially introducing noise into the training process. Second, the limited size of these datasets makes it difficult to train robust neural models without overfitting. Lastly, the cultural and linguistic nuances of these languages may not be fully

---

[1]https://ufal.mff.cuni.cz/malayalam-visual-genome

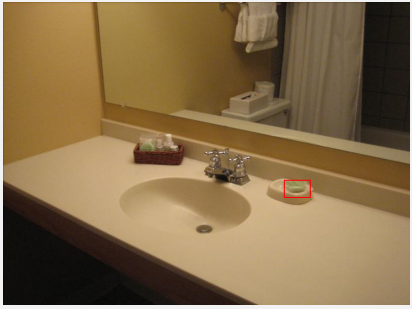

**Source English Caption:**
soap is in the dish
**Target Hindi Caption:**
साबुन पकवान में है

Table 1: Illustration of a translation ambiguity in the dataset. The English caption for the bounded region **"soap is in the dish"** is mistranslated into Hindi as "साबुन पकवान में है" (**"soap is in the food"**). The Hindi word "पकवान" typically means **"food dish"**, whereas the visual context clearly indicates that **"dish"** refers to a **soap holder**. This example also highlights the importance of visual context in resolving lexical ambiguities in multimodal translation tasks.

captured in direct translations (Hershcovich et al., 2022) of English captions.

Existing approaches (Calixto and Liu, 2017) to MMT often involve training complex neural architectures on large parallel corpora with paired images. However, such methods may struggle with the limited and potentially noisy data available for low-resource languages. Moreover, these approaches often fail to leverage the rich semantic understanding capabilities of recent large language models (LLMs) along with their proficiency in visual (Liu et al., 2024; Radford et al., 2021) and multilingual (Üstün et al., 2024; Workshop et al., 2022; Dubey et al., 2024; Touvron et al., 2023; Jiang et al., 2023) understanding gained through training on large corpora of data across domains.

In this paper, we propose a pipeline that addresses these challenges by leveraging multi-modal LLMs, specifically GPT-4o[2] and Claude 3.5 Sonnet[3], to enhance cross-lingual image captioning. Our approach uses instruction-tuned prompting to generate rich, contextual conversations about images using the English captions as context along with the image. These synthetic conversations, comprising detailed descriptions, simple question-answer (QA) pairs, and complex reasoning question-answer pairs, are then translated into the target languages and used to inform the final caption generation in the target language. This method allows us to:

- Mitigate the impact of limited and noisy training data by generating synthetic, high-quality contextual information.

- Leverage the advanced reasoning capabilities of LLMs to provide culturally and linguistically nuanced translations.

- Explore the balance between source fidelity and enhanced description through a weighted prompting strategy.

The main contributions of our work are:

- A pipeline for low-resource MMT that requires no traditional training or fine-tuning.

- A weighted prompting mechanism that calibrates between source caption fidelity and LLM-generated contextual information, facilitating a nuanced balance of translation accuracy, caption diversity, and exhaustive visual description coverage.

- A framework for dataset enrichment through the generation of detailed descriptions and complex reasoning QA pairs to augment existing multimodal datasets.

- Empirical analysis of LLMs' capabilities in direct translation and target language summarization, providing insights into their potential for low-resource languages.

## 2 Dataset

We utilized only the datasets specified by the organizers for the related tasks. However, our use of the GPT-4o and Claude 3.5 Sonnet models places our submissions in the unconstrained track. The provided datasets contain captions in English and the target language, describing rectangular regions in images of various scenes. The task involves generating captions in the target language using either the text, the image, or both. Across all languages, the training set consists of 29,000 examples. The dataset is complemented by three test sets: development (D-Test), evaluation (EV-Test), and challenge (CH-Test). Our submissions were evaluated on the EV-Test and CH-Test sets, which contain approximately 1,600 and 1,400 examples, respectively. The development set comprises around 1,000 examples.

---

[2]gpt-4o-2024-08-06: https://platform.openai.com/docs/models/gpt-4o.

[3]Claude 3.5 Sonnet: https://www.anthropic.com/api.

Table 2 shows the parallel corpus statistics across the various languages. Table 3 shows that data sources of datasets for each task.

## 3 Methodology

The overall pipeline of our approach is shown in Figure 1.

### 3.1 Preprocessing

For the text data, all utterances are converted to lowercase, and punctuation is removed. The dataset includes images of complete scenes along with the coordinates of the bottom-left corner and the dimensions of the rectangular region corresponding to each caption. This information is used to crop the relevant rectangular regions from the images. Since these images are later used as part of prompts for LLMs, base64 encodings of all images in the EV and CH sets are generated.

### 3.2 Multi-Model Context Generation in English with a Fusion approach

In this step , we leverage the capabilities of two large language models (LLMs) - GPT-4o and Claude 3.5 Sonnet - to generate rich, contextual conversations about the input image and its associated English caption. This process involves two key stages: individual LLM processing and conversation fusion.

We separately prompt GPT-4o and Claude 3.5 using Prompt-1 from Table 4. The format of the conversation and prompt design is inspired by an example prompt from Liu et al. (2024). Both models are given the same input: the cropped image and its English caption. This parallel processing allows us to capitalize on the unique strengths of each model.

After obtaining separate conversations from GPT-4o and Claude 3.5, we employ a fusion

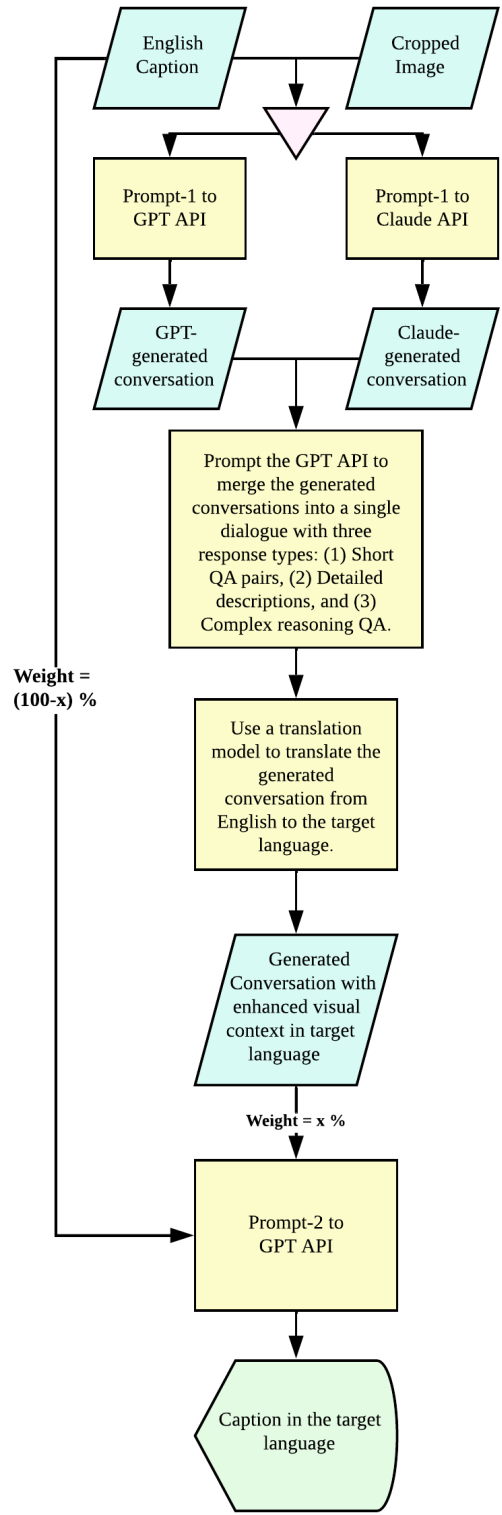

Figure 1: Overview of the system pipeline. Prompt-1, detailed in Table 4, and Prompt-2, detailed in Table 5, respectively. The weight x% must be specified during querying.

model, implemented using the GPT-4o API,

| Set | Sentences | English | Hindi | Malayalam | Bengali | Hausa |
|---|---|---|---|---|---|---|
| **Train** | 28930 | 143164 | 145448 | 107126 | 113978 | 140981 |
| **D-Test** | 998 | 4922 | 4978 | 3619 | 3936 | 4857 |
| **E-Test** | 1595 | 7853 | 7852 | 5689 | 6408 | 7736 |
| **C-Test** | 1400 | 8186 | 8639 | 6044 | 6657 | 8752 |

Table 2: Parallel Corpus Token statistics for each dataset split across different languages.

| Task | Source |
|---|---|
| English→Hindi | HindiVisualGenome1.1 (Parida et al., 2019) |
| English→Malayalam | MalayalamVisualGenome1.0[4] |
| English→Bengali | BengaliVisualGenome1.0 (Sen et al., 2022) |
| English→Hausa | HausaVisualGenome1.0 (Abdulmumin et al., 2022) |

Table 3: Tasks and their corresponding dataset sources.

to integrate these conversations. This fusion process is designed to create a single, coherent dialogue that encompasses three response types: **(1) Short QA pairs, (2) Detailed descriptions, and (3) Complex reasoning QA**. The prompt has been designed to generate such types of responses to ensure that the responses can capture the context effectively. The fusion prompt is carefully crafted to ensure that the final conversation retains the most relevant and insightful elements from both initial conversations.

This fusion approach is aimed at using complementary strengths of the models for error mitigation and rich context consolidation. Recent work has shown the advantage of such fusion and ensembling approaches in mitigating hallucination across tasks such as machine translation, definition modeling, and paraphrase generation (Mehta et al., 2024). Additionally, such an approach has a flexibility advantage as it allows for future integration of additional LLMs or specialized models.

### 3.3 Translation of Context to Target Languages

This stage involves translating the rich contextual information generated in English to the target languages. This step is essential for preserving the nuanced understanding developed in the previous stages while adapting it to the linguistic and cultural context of each target language. For the translation of the generated conversations, we employ different approaches based on the target language:

- **Hindi, Bengali, and Malayalam:** For these Indic languages, we utilize the IndicTrans2 (Gala et al., 2023) model. This state-of-the-art translation model is specifically designed for Indian languages and has demonstrated strong performance in multilingual translation tasks.

- **Hausa**: Due to the limited availability of specialized translation models for Hausa, we leverage the GPT-4o's translation capabilities.

### 3.4 Weighted Prompt-Based Caption Generation

The final stage of our pipeline employs a weighted prompting strategy to generate the target language caption, balancing fidelity to the original English caption with the rich contextual information derived from our LLM-generated conversations. We utilize GPT-4o API for this crucial step, employing a carefully crafted prompt (Prompt-2, detailed in Table 5) that takes two primary inputs along with the weight value:

1. The original English caption (weight: 100-x%)

2. The translated conversation in the target language (weight: x%)

The weighting mechanism allows us to control the influence of each input on the final caption. This approach offers flexibility in balancing between direct translation fidelity and contextual enrichment. For our submissions we provide equal weightage to the given English caption and the generated conversation in the target language.

## 4 Results

The BLEU score (Papineni et al., 2002) serves as the primary metric for evaluating model performance on the leaderboard, complemented

You are an AI visual assistant, and you are seeing a single image. You are provided with an English caption describing the image you are looking at. Answer all questions as if you are seeing the image. Design a conversation between you and a person asking about this photo. The answers should be in a tone that a visual AI assistant would use while seeing the image and answering the questions. Ask diverse questions and give corresponding answers.

Include questions asking about the visual content of the image, including the object types, counting the objects, object actions, object locations, relative positions between objects, etc. Only include questions that have definite answers:

(1) one can see the content in the image that the question asks about and can answer confidently;

(2) one can determine confidently from the image that it is not in the image. Do not ask any question that cannot be answered confidently.

Also, include complex questions that are relevant to the content in the image, for example, asking about background knowledge of the objects in the image, asking to discuss events happening in the image, etc. Again, do not ask about uncertain details. Provide detailed answers when answering complex questions. For example, give detailed examples or reasoning steps to make the content more convincing and well-organized. You can include multiple paragraphs if necessary. You must include a question-answer pair asking about the number of words in the English caption.

I am giving you an example of a conversation too, so that you can follow a format in the conversation you generate.

Example Image: {example_image_link}

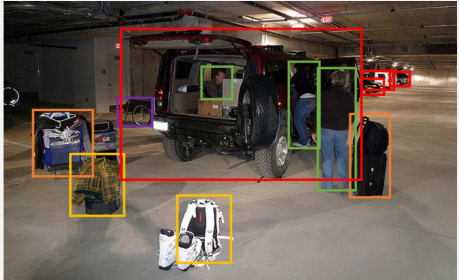

Associated Caption in English: A group of people standing outside of a black SUV with various luggage.

Associated Conversation in English:

Response type 1: conversation

Question: What type of vehicle is featured in the image?

Answer: The image features a black sport utility vehicle (SUV).

Question: Where is the vehicle parked?

Answer: The vehicle is parked in an underground parking area, likely in a public garage.

Question: What are the people in the image doing?

Answer: The people are trying to fit all of their luggage into the SUV, likely preparing for a trip.

Response type 2: detailed description

The image shows an underground parking area with a black SUV. Three people are packing luggage into the vehicle, with one person on the left, one in the middle, and one on the right. Scattered around the SUV are two backpacks—one near the left rear wheel, another on the right—and two suitcases, one beside the car and another near the center of the lot. A bicycle is visible on the left. Other cars are parked nearby: one directly behind, one slightly to the left, and another to the right.

Response type 3: complex reasoning

Question: What challenges do these people face?

Answer: A group of people stands outside a black SUV in a parking area, surrounded by suitcases and backpacks. They face the challenge of fitting all their luggage into the vehicle, indicating they have a lot to pack. They need to arrange the luggage efficiently to ensure everything fits and consider the comfort of passengers and the driver's visibility during the trip.

As you can see from the example, the generated text should have 3 response types: A short question-and-answer type conversation, a detailed description, and a question-answer pair for complex reasoning. Now, here is the caption and the image I want you to generate the conversation for: {english_caption}, {image_link}. Do not hallucinate!

Table 4: Prompt-1: This illustrates the prompt construction process for generating the conversation response using an AI visual assistant. The code snippet generates a prompt for GPT-4 to create a conversation based on an image and caption. The conversation includes three response types: a short Q&A, a detailed description, and complex reasoning. The generated response is capped to a maximum of 1024 tokens for this step.

> **Task:** Generate a caption in {target_language} for an image, balancing information from the English caption and additional context.
> **Context:**
> 1. English caption (Weight: {english_weight}%): {english_caption}
> 2. Additional context in {target_language} (Weight: {context_weight}%): {translated_conversation}
> **Requirements:**
> 1. Create a caption that combines information from both sources according to their weights:
>  - {english_weight}% of the information should come from the English caption's structure and details.
>  - {context_weight}% of the information can be enriched or modified based on the additional context.
> 2. Maintain a level of detail appropriate to the weighted combination of sources.
> 3. Use natural, fluent {target_language} appropriate for image captions.
> 4. Ensure the final caption length is similar to the English caption, adjusting slightly if necessary for language differences.
> **Output:**
> Provide ONLY the {target_language} caption. The output must be exclusively in {target_language} without any additional text or explanations. **Do not hallucinate!**

Table 5: Prompt-2: This prompt guides the generation of a caption in the target language by balancing the original English caption with additional context, weighted according to specified ratios. The generated caption should reflect the natural fluency of the target language and match the original caption's length. The generated response is capped to a maximum of 300 tokens for this step.

by the RIBES metric (Isozaki et al., 2010) for a more comprehensive assessment. Table 6 presents our results across all language pairs and datasets. Notably, our approach achieves significant success in the English-Hausa task, securing the 1st position on the Challenge Leaderboard and the 2nd position on the Evaluation Leaderboard. For the English-Hindi task, we demonstrate competitive performance, obtaining strong BLEU and RIBES scores on both the evaluation and challenge sets. Our method, which requires no training, yields competitive results across all 8 submissions, underscoring its effectiveness and versatility. However, we observe relatively limited performance in the Bengali, Hausa, and Malayalam tasks. This performance discrepancy likely reflects the current limitations of state-of-the-art LLMs, whose training data may not adequately represent low-resource languages and their unique semantic characteristics.

### 4.1 Weight Tuning Analysis and Performance Metrics

To thoroughly evaluate our weighted prompt-based approach, we conducted extensive experiments across different weight combinations for each target language. This section presents our findings and analyzes the impact of various weight distributions on caption quality and semantic preservation. We employed three primary metrics to assess the performance of our model under different weight configurations:

| System and Task | BLEU | RIBES | Position |
|---|---|---|---|
| **English→Hindi** | | | |
| MMEVMM22en-hi | 29.70 | 0.725 | 5th |
| MMCHMM22en-hi | 37.90 | 0.796 | 3rd |
| **English→Malayalam** | | | |
| MMEVMM22en-ml | 15.10 | 0.411 | 4th |
| MMCHMM22en-ml | 13.60 | 0.428 | 3rd |
| **English→Bengali** | | | |
| MMEVMM22en-bn | 22.10 | 0.575 | 5th |
| MMCHMM22en-bn | 21.70 | 0.644 | 4th |
| **English→Hausa** | | | |
| MMEVMM22en-ha | 17.70 | 0.580 | 2nd |
| MMCHMM22en-ha | 21.10 | 0.637 | 1st |

Table 6: Summary of results for various English-to-target language multimodal tasks. The table shows BLEU, RIBES scores, and positions for different tasks.

| Weight | BLEU | Sem. Sim. | Norm. Sem. |
|---|---|---|---|
| 0 | 28.56 | 0.9238 | 0.9738 |
| 10 | 23.10 | 0.8978 | 0.9715 |
| 20 | 21.46 | 0.8866 | 0.9726 |
| 30 | 18.73 | 0.8504 | 0.9723 |
| 40 | 16.52 | 0.8381 | 0.9800 |
| 50 | 16.51 | 0.8503 | 0.9747 |
| 60 | 13.71 | 0.7972 | 0.9850 |
| 70 | 13.01 | 0.8103 | 0.9822 |
| 80 | 11.57 | 0.8066 | 0.9784 |
| 90 | 11.26 | 0.8158 | 0.9761 |
| 100 | 10.22 | 0.7893 | 0.9722 |

Table 7: Results for Hindi (hi) with varying context weights. The table shows Average BLEU, Semantic Similarity (hi-hi), and Normalized Semantic Similarity (en-hi).

| Weight | BLEU  | Sem. Sim. | Norm. Sem. |
|--------|-------|-----------|------------|
| 0      | 12.24 | 0.7575    | 0.9380     |
| 10     | 16.47 | 0.7379    | 0.9426     |
| 20     | 11.67 | 0.6992    | 0.9215     |
| 30     | 14.29 | 0.6189    | 0.9437     |
| 40     | 10.06 | 0.5780    | 0.9532     |
| 50     | 8.83  | 0.5998    | 0.9611     |
| 60     | 8.43  | 0.5446    | 0.9570     |
| 70     | 7.10  | 0.5091    | 0.9559     |
| 80     | 8.50  | 0.5460    | 0.9617     |
| 90     | 8.81  | 0.5185    | 0.9583     |
| 100    | 6.43  | 0.5091    | 0.9474     |

Table 8: Results for Malayalam (ml) with varying context weights. The table shows Average BLEU, Semantic Similarity (ml-ml), and Normalized Semantic Similarity (en-ml).

| Weight | BLEU  | Sem. Sim. | Norm. Sem. |
|--------|-------|-----------|------------|
| 0      | 24.23 | 0.9348    | 0.9617     |
| 10     | 20.07 | 0.9266    | 0.9699     |
| 20     | 17.21 | 0.9072    | 0.9677     |
| 30     | 14.53 | 0.8957    | 0.9811     |
| 40     | 14.98 | 0.8607    | 0.9681     |
| 50     | 13.92 | 0.8670    | 0.9653     |
| 60     | 10.11 | 0.8319    | 0.9829     |
| 70     | 10.29 | 0.8380    | 0.9925     |
| 80     | 10.23 | 0.8545    | 0.9827     |
| 90     | 11.24 | 0.8498    | 0.9789     |
| 100    | 9.35  | 0.8486    | 0.9653     |

Table 9: Results for Bengali (bn) with varying context weights. The table shows Average BLEU, Semantic Similarity (bn-bn), and Normalized Semantic Similarity (en-bn).

| Weight | BLEU  | Sem. Sim. | Norm. Sem. |
|--------|-------|-----------|------------|
| 0      | 51.58 | 0.6187    | 0.9126     |
| 10     | 46.38 | 0.5599    | 0.9629     |
| 20     | 45.36 | 0.5358    | 0.9505     |
| 30     | 42.23 | 0.5031    | 0.9598     |
| 40     | 37.58 | 0.4829    | 0.9701     |
| 50     | 36.14 | 0.4712    | 0.9556     |
| 60     | 30.48 | 0.4366    | 0.9667     |
| 70     | 28.68 | 0.4344    | 0.9708     |
| 80     | 29.20 | 0.4407    | 0.9757     |
| 90     | 29.41 | 0.4361    | 0.9642     |
| 100    | 32.01 | 0.4442    | 0.9738     |

Table 10: Results for Hausa (ha) with varying context weights. The table shows Average BLEU, Semantic Similarity (ha-ha), and Normalized Semantic Similarity (en-ha).

- **BLEU Score:** Bilingual Evaluation Understudy (BLEU) (Papineni et al., 2002) measures the similarity between the generated caption and the reference caption. It provides a quantitative measure of translation quality.

- **Semantic Similarity (Sem. Sim.):** We use cosine similarity between sentence embeddings to measure the semantic closeness of the generated caption to the reference caption in the target language. This metric is calculated using the SentenceTransformer's SentenceBERT (Reimers and Gurevych, 2019) model 'distiluse-base-multilingual-cased-v1', which provides multilingual sentence embeddings.

- **Normalized Semantic Similarity (Norm. Sem.):** This metric compares the semantic similarity of the generated caption to the English source with that of the reference translation to the English source as the ratio of the two. It helps assess how well the generated caption preserves the meaning of the original English caption relative to the reference translation.

To ensure a comprehensive evaluation of our approach across various scenarios, we conducted an in-depth analysis on a diverse subset of the corpus. This subset comprises 250 image-caption pairs, randomly selected from the training, development, evaluation, and challenge sets. Tables 7, 8, 9, and 10 present the results for Hindi, Malayalam, Bengali, and Hausa, respectively. The 'Weight' column represents the percentage weight given to the translated conversation, with the complementary weight assigned to the original English caption. The reported metrics are averaged over all 250 data points for each weight configuration.

Generally, BLEU scores decrease as more weight is given to the translated conversation. This suggests that higher weights on the original caption produce translations more closely aligned with the reference.

The semantic similarity between the generated and reference captions in the target language tends to decrease with increasing weight on the translated conversation. This indicates that while the generated captions may become more descriptive, they may deviate from the reference in terms of semantic content.

Interestingly, the normalized semantic similarity remains relatively stable across weight distributions. This suggests that our approach consistently preserves the semantic relation-ship between the English source and the generated caption, regardless of the weight distribution.

While the highest BLEU scores are generally achieved with lower weights on the translated conversation, there's a trade-off between BLEU score and the richness of the generated caption. A balanced approach (e.g., 50-50 weighting) often provides a good compromise between translation accuracy and contextual enrichment.

Notably, setting a 100% weight for the original English caption allows us to evaluate GPT-4's zero-shot cross-lingual transfer capabilities in direct translation tasks from English to Hindi, Bengali, Malayalam, and Hausa. Conversely, assigning a 100% weight to the additional context—which consists of the translated LLM-generated conversation in the target language—enables us to assess the model's abstractive summarization abilities in these non-English languages. This analysis provides insights into the models' multilingual competence and their capacity for language understanding and generation across diverse linguistic contexts.

## 5 Conclusion

Key strengths of our approach include its training-free nature, which avoids propagating errors from potentially flawed datasets, and its flexibility in balancing source fidelity with enhanced descriptiveness through a weighted prompting strategy. The method's multilingual capability and rich context generation offer promising avenues for dataset enrichment and improvement in low-resource languages. In Section 4.1, we demonstrated how our weighted prompting strategy serves as a probe for assessing LLMs' capabilities in zero-shot cross-lingual transfer for direct translation tasks, as well as their abstractive summarization abilities in low-resource target languages. However, we acknowledge limitations such as reliance on LLM APIs, potential for hallucination, and computational intensity. The challenge of evaluating enhanced descriptions with traditional metrics like BLEU also highlights the need for more comprehensive evaluation methods. Future work should focus on:

- Conducting human evaluations to better assess caption quality and appropriateness.
- Analyzing specific cases of significant improvements or detractions from original captions.
- Exploring applications in dataset error correction and enhancement.
- Investigating performance across diverse image types and caption complexities.

By addressing these areas, we aim to further refine and expand the capabilities of our approach, potentially leading to more robust and versatile multimodal translation systems. This work represents a step towards bridging the gap between high-resource and low-resource languages in multimodal machine translation, opening new possibilities for cross-lingual image understanding and dataset enrichment.